\documentclass[runningheads]{llncs}

\usepackage{graphicx}
\usepackage{hyperref}
\usepackage{xcolor}

\usepackage[title]{appendix}
\usepackage{url}
\usepackage{booktabs}
\usepackage{amsmath}
\usepackage{amsfonts}
\usepackage{todonotes}
\usepackage{mathtools}

\DeclareMathOperator*{\argmax}{arg\,max}

\usepackage{xspace}
\def\systemname#1{\textsf{#1}\xspace}
\newcommand{\lc}{\systemname{leanCoP}}
\newcommand{\rlc}{\systemname{rlCoP}}
\newcommand{\plc}{\systemname{plCoP}}
\newcommand{\gc}{\systemname{graphCoP}}

\begin{document}

\title{The Role of Entropy\\ in Guiding a Connection Prover}

\author{
  Zsolt Zombori\inst{1,2}
  \and Josef Urban\inst{3}
  \and Miroslav Ol\v{s}\'{a}k\inst{4}
}

\authorrunning{Zombori, Urban, Ol\v{s}\'{a}k}
\titlerunning{Role of Entropy in Guiding a Prover}

\institute{
  Alfr\'{e}d R\'{e}nyi Institute of Mathematics, Budapest
  \and E{\"o}tv{\"o}s Lor\'{a}nd University, Budapest
  \and Czech Technical University in Prague
  \and University of Innsbruck  
}

\maketitle

\begin{abstract}
In this work we study how to learn good algorithms for selecting reasoning steps in theorem proving.
We explore this in  the connection tableau calculus implemented by \lc where the partial tableau provides a clean and compact notion of a \emph{state} to which a limited number of inferences can be applied.
We start by incorporating a state-of-the-art learning algorithm -- a
graph neural network (GNN) -- into the \plc theorem prover. Then we
use it to observe the system's behavior in a reinforcement learning
setting, i.e., when learning 
inference
guidance from successful Monte-Carlo tree searches on many problems.
Despite its better pattern matching capability, the GNN 
initially performs worse than a simpler previously used learning algorithm.
We observe that the simpler algorithm is less confident, i.e., its recommendations have higher entropy.
This leads us to explore how the entropy of the inference selection 
implemented via the neural network 
influences the proof search. This is related to research in human decision-making under uncertainty, and in particular the \emph{probability matching} theory.
Our main result shows that a proper entropy regularization, i.e., training the GNN not to be overconfident, 
greatly improves  \plc's performance on a large mathematical corpus.
\end{abstract}

\section{Introduction}

Automated Theorem Proving (ATP) and Interactive Theorem Proving (ITP) 
are today increasingly benefiting from combinations with Machine
Learning (ML) methods~\cite{JosefsFinalReport}.
A number of  learning-based inference guiding methods have been
developed recently, starting with the \lc \cite{leancop,DBLP:conf/cade/Otten08} style connection tableaux setting~\cite{UrbanVS11,KaliszykU15,mfckju-jar21,rlcop,plcop,EasyChair:4433,prop_invariant_embedding},
later expanding into the E prover's \cite{Sch02-AICOMM,Schulz13} and Vampire's  \cite{Vampire}
 superposition setting~\cite{JakubuvU17a,deep_guidance,ChvalovskyJ0U19,JakubuvCOP0U20,SudaCADE21},
 and HOL's \cite{Gordon:1993:ITH,Harrison96,SlindN08}, Coq's \cite{coq} and Isabelle's~\cite{WenzelPN08} tactical settings~\cite{tgckju-lpar17,NagashimaH18,HuangDSS19,YangD19,BansalLRSW19,blaauwbroek2020tactic,tgckjurkmn-jar21}.

 The connection tableau calculus as implemented by \lc is a very good
 framework for studying combinations of ML and ATP
 methods~\cite{Bibel17}.  \lc has a compact Prolog implementation that
 is both easy to modify and surprisingly efficient. At the same time,
 unlike in the superposition and tactical setting, the partial tableau
 provides a clean and compact notion of a \emph{state} to which a
 limited number of inferences (\emph{actions}) can be applied. This has recently allowed the
 first experiments with AlphaZero \cite{alphazero} style Monte-Carlo tree search (MCTS) \cite{mfckju-jar21}
 and Reinforcement
 Learning (RL) \cite{Sutton1998} of theorem proving in the \rlc \cite{rlcop}, \gc \cite{prop_invariant_embedding} and \plc \cite{plcop} systems.

In this work, we start by  extending \plc with a state-of-the-art
learning algorithm –- a graph neural network (GNN)   
\cite{prop_invariant_embedding} -- which was designed for processing
logical formulae and exhibits several useful invariance properties,
namely invariance under renaming of symbols, negation and reordering
of clauses and literals.
Despite its better pattern matching capability, the GNN initially performs worse than a
simpler previously used learning algorithm based on gradient boosted trees (XGBoost~\cite{xgboost}). 
We observe that the simpler
algorithm is less confident about the inferences that should be applied to the proof states, i.e., its recommendations have higher entropy, leading to greater exploration of different inferences.

This leads us to analyze how the entropy of the inference selection implemented via the neural network influences the proof search.
We try increasingly high penalties for overconfidence (low entropy) during the training of the GNN, using an approach called
Maximum Entropy Reinforcement Learning~\cite{maxent_first}. 
For this, we need to be able to compare the entropy of proof states with different numbers of possible inferences (actions).
We do that by introducing \emph{normalized entropy}, which allows for comparing
discrete distributions of different lengths.
We make a rather surprising discovery that replacing the particular trained predictors
by arbitrary (random) but entropy-normalized predictors that respect the inference ordering 
yields 
only slightly weaker ATP
performance. This suggests that the correct ordering of possible inferences according to their utility in a given state plus the
right amount of entropy capture most of the benefits of the learned 
guidance.
In summary, our contributions are:

\begin{enumerate}
\item We integrate a fast logic-aware graph neural network into the \plc system, allowing its use for guiding the choice of inferences (policy) and for estimating the provability of a partial connection tableau (value).

\item We adapt the graph construction algorithm to support the paramodulation inferences used in \plc.

\item We analyze the entropy of the policy and its role in \plc's performance.

\item We show that random policies with the right ordering and normalized entropy perform already quite well.

\item We do several smaller and larger evaluations over the standard corpus extracted from the Mizar Mathematical Library (MML) and show
that the best entropy regularized GNN 
greatly improves over other learning-guided connection tableaux
systems. In particular, we report $17.4\%$ improvement on the Mizar40
evaluation set over \rlc, the best previously published result.

\end{enumerate}

The rest of the paper is structured as
follows. Section~\ref{background} introduces in more detail the
necessary background such as neural guidance of provers, the \lc
setting, reinforcement learning and Monte-Carlo tree search, and
Maximum Entropy learning. Section~\ref{sec:maxent_atp}  discusses in more depth
the use of Maximum Entropy learning in guiding MCTS-based theorem
proving. Section~\ref{implementation} describes our new implementation
and Section~\ref{sec:experiments} experimentally evaluates the methods.

\section{Background and Related Work}
\label{background}

\subsection{Neural Feature Extraction for Guiding Theorem Provers}
\label{neural}
Learning based ATP systems have for a long time explored suitable
characterizations of mathematical objects, leading to solutions that
process text directly (e.g. \cite{deepmath,deep_guidance,BansalLRSW19}) and
solutions that rely on manually engineered features
(e.g. \cite{ckjujv-ijcai15,JakubuvCOP0U20}). Graph
neural networks (GNN) \cite{gnn} provide an alternative to both
approaches: the graph representation allows for retaining the syntactic
structure of mathematical objects, while also allowing for end-to-end (i.e., involving no manually designed features)
training. However, improving over learning based on manual feature extraction
has proven to be challenging with GNNs, especially in real
time, as noted in several works (e.g. \cite{ChvalovskyJ0U19,Crouse2019ADR}).
Usually it required high level of technical engineering. The GNN
presented in \cite{prop_invariant_embedding} was designed to preserve
many useful invariance properties of logical formulae and has
demonstrated impressive improvement in guiding the \lc connection calculus
compared with gradient boosted trees. We refer to this system as \gc.

\subsection{Systems Guiding the \lc Theorem Prover}
\lc~\cite{leancop} is a compact theorem prover for first-order logic,
implementing connection tableau search. The proof search starts with a
\emph{start clause} as a \emph{goal} and proceeds by building a
connection tableau by applying \emph{extension steps} and
\emph{reduction steps}.  \lc uses iterative deepening to ensure
completeness.

A series of learning systems guiding the \lc connection calculus have been developed recently. Of
these, we highlight three that use roughly the same reinforcement learning setup:
\rlc~\cite{rlcop}, \plc~\cite{plcop} and
\gc~\cite{prop_invariant_embedding}. These systems search for proofs
using Monte Carlo Tree Search~\cite{mcts} and they train the \emph{value} (the proof state quality) and
the \emph{policy} (the inference quality in a proof state) functions similarly to systems like AlphaZero \cite{alphazero,thinking_fast_and_slow}.
\rlc and \plc use manually developed features~\cite{ckjujv-ijcai15}
and gradient boosted trees (XGBoost~\cite{xgboost}) for learning while
\gc employs a GNN for end-to-end feature extraction and learning.
This graph neural network was designed for processing mathematical
formulae and has several useful invariance properties: the graph
structure is invariant under renaming of symbols, negation and reordering of
clauses and literals. In this paper, we incorporate the GNN of
\cite{prop_invariant_embedding} into \plc.

The \plc system extends \lc with paramodulation steps that can handle equality
predicates more efficiently. Let $t|_p$ denote the subterm of $t$ at
position $p$ and $t[u]_p$ denote the term obtained after replacing in
$t$ at position $p$ by term $u$. Given a goal $G$ and 
an input clause~\footnote{Assuming Disjunctive Normal Form.} $\{X \neq Y, B\}$, 
such that, for some position $p$ there is a
substitution $\sigma$ such that $G|_p\sigma = X\sigma$, the paramodulation
step changes $G$ to $\{G[Y]_p\sigma, B \sigma\}$.  Rewriting is
allowed in both directions, i.e., the roles of $X$ and $Y$ can be
switched.

\subsection{Reinforcement Learning (RL)}
Reinforcement learning (RL) \cite{Sutton1998} aims to find the optimal
behaviour in an environment defined as a Markov Decision Process
(MDP).  An $\mbox{MDP}(\mathcal{S}, \mathcal{A}, \mathcal{R},
\mathcal{P}, \gamma)$ describes a dynamic process and consists of the
following components: $\mathcal{S}$ is the set of possible states,
$\mathcal{A}$ is the set of possible actions,
$\mathcal{R}:(\mathcal{S} \times \mathcal{A}) \rightarrow \mathbb{R}$
is a reward function, $\mathcal{P}: (\mathcal{S} \times \mathcal{A})
\rightarrow \mathcal{S}$ is the state transition function and $\gamma$
is the discount factor. We assume that an agent interacts with this
MDP, generating sequences of $(s_t, a_t, r_t)$ state-action-reward
tuples, called \emph{trajectories}. The agent is equipped with a
\emph{policy} function $\pi: \mathcal{S} \rightarrow \mathcal{A}$
which determines which action it selects in a particular state. The
policy is often stochastic, i.e., it defines a probability
distribution over inferences that are possible in a given state. The
aim of the agent is to maximize its total accumulated reward $\sum_{t
  \geq 0} \gamma^{t} r_t$.  Several components, such as the reward and
transition functions, can be stochastic, in which case the aim of the
agent it to find the policy $\pi^*$ that maximizes its cumulative
expected reward, where future rewards are discounted with the $\gamma$
discount factor:

$$ \pi^* = \argmax_{\pi} \mathbb{E} \left\lbrack \sum\limits_{t \geq 0}
\gamma^{t} r_t | \pi \right\rbrack$$

\subsection{Monte Carlo Tree Search (MCTS)}
\label{sec:mcts}
MCTS is a simple RL algorithm, which builds a search tree whose nodes
are states and where edges represent actions. The aim of the search
algorithm is to find trajectories (branches in the search tree) that
yield high accumulated rewards. The search starts from a single root node,
and new nodes are added iteratively. In each node $i$, we maintain the
number of visits $n_i$, the total reward $r_i$, and the prior
probability (estimated typically by a trained predictor) $p_i$ of all its possible successors (in our case produced by the possible inferences).
Each iteration, also called
\emph{playout}, involves the addition of a new leaf node.

The
policy $\pi$ used for selecting the actions of the playout is based on
the standard UCT~\cite{uct} (Upper Confidence Trees) formula (\ref{eq:uct}): in each state we
select the action with the maximal UCT value in the successor
state. Once a new leaf state is created, we observe its reward and
update its ancestors: visit counts are increased by 1 and rewards are
increased by the reward of the leaf.

\begin{equation}\label{eq:uct}
  \mbox{UCT}(i) = \frac{r_i}{n_i} + cp \cdot p_i \cdot \sqrt{\frac{ln N}{n_i}}
\end{equation}

\noindent In~(\ref{eq:uct}), $N$ is the number of visits of the
parent, and $cp$ is a parameter that determines the balance between
nodes with high reward (exploitation) and rarely visited nodes
(exploration).
In \cite{thinking_fast_and_slow,alphazero} MCTS is augmented with two learned
functions. The \emph{value} function estimates the accumulated reward
obtainable from a state, and the leaf nodes are initialized with this value
estimates. 
The second function predicts
the prior probability of state-action pairs, which is usually
referred to as the \emph{policy}, with a slight abuse of
terminology. When it can lead to confusion, we refer to this policy as $\pi^M$.

The \plc, \rlc and \gc systems use the MaLARea/DAgger~\cite{US+08,dagger}
meta-learning algorithm to learn the policy and value functions. They
interleave ATP runs based on the current policy and value (\emph{data
  collection phase}) with a \emph{training phase}, in which these
functions are updated to fit the collected data. Such iterative
interleaving of proving and learning has  been used successfully
in ATP systems such as MaLARea~\cite{US+08} and
ENIGMA~\cite{JakubuvU19}.

During the proof search we build a Monte Carlo tree for each training
problem. Its nodes are the proof states (partial tableaux), and the
edges represent inferences. Note that the Monte Carlo tree is
  thus different from the tableau trees.  A branch of this Monte
Carlo tree leading to a node with a closed tableau is a valid
proof. Initially, the three \lc-based systems use somewhat different
heuristic value and policy functions, later to be replaced with the
learned guidance.  To enforce deeper exploration, we perform a
\emph{bigstep} after a fixed number of playouts: the starting node of
exploration is moved one level down towards the child with the highest
value (called the \emph{bigstep node}). Later MCTS steps thus only
extend the subtree under the bigstep node. This in practice means no
backtracking of the bigsteps, which in turn involves giving up
completeness.

\subsection{Maximum Entropy Reinforcement Learning}
\label{sec:maxent}
When training the policy, directly maximizing the expected utility on the action sequences
observed by an RL agent (i.e., the training examples) can lead to instability.
The policy can get stuck in local minima and become overconfident, preventing it from
exploring the search space sufficiently when necessary to make good decisions. This
has motivated using stochastic policies and several \emph{regularization} (i.e., encouraging generality)
techniques that ensure that all actions have a chance of being selected.
Another motivation for properly regularized stochastic policy learning
comes from experiments on humans and animals, suggesting that
biological agents do not deterministically select the action with the
greatest expected utility \cite{probability_matching}: instead they
randomly select actions with probability proportional to the expected
utility, called \emph{probability matching}. Consequently, action
sequences that generate similar rewards tend to be similarly probable,
i.e., we avoid making strong commitments whenever it is possible.
Maximum Entropy Reinforcement Learning (MaxEnt RL), achieves
probability matching by adding an entropy term to the loss founction when the policy is trained:

$$ \pi^* = \argmax_{\pi} \mathbb{E} \lbrack \sum\limits_{t \geq 0}
\gamma^{t} r_t + \alpha H_{\pi}[a|s_t] | \pi \rbrack$$

\noindent where  $H_{\pi}[a|s_t]$ is the Shannon entropy of the
probability distribution over valid actions in state $s_t$:

$$H[p] = - \sum_{i=1}^n p_i \log(p_i)$$

\noindent and $\alpha$ is the entropy coefficient.
This means that the training of the policy will be maximizing a weighted sum of the (discounted) rewards
and of the entropy of the resulting distribution, thus discouraging overconfidence.
The entropy term in
the objective was first used in ~\cite{maxent_first} and since then
its benefit has been empirically demonstrated in several domains. It
is particularly useful in dynamic environments, where some uncertainty
remains, irrespective of the amount of exploration. 

\subsection{Kullback-Leibler divergence}
Shannon's entropy measures the uniformity of a single
distribution. However, when contrasting different policies, we will
need a measure to compare different distributions. For this, one of
the most widely used options is the Kullback-Leibler (KL) divergence,
also called \emph{relative entropy}, which is a measure of how one
probability distribution differs from a given reference
distribution. For a discrete target distribution $Q$ and a reference
distribution $P$, the KL divergence is defined as:

$$KL(P \lVert Q) = \sum_x P(x) \log \frac{P(x)}{Q(x)}$$

\noindent This measure is zero exactly when $P$ and $Q$ are identical,
otherwise it is positive. It can be infinite if there is some $x$ such
that $P(x) > 0$ and $Q(x) = 0$. A small $KL(P \lVert Q)$ means that
the two distributions are similar on the domain where most of the
probability mass of $P$ lies. Note that $KL(P \lVert Q) \neq KL(Q
\lVert P)$. For example, consider the following distributions:

\begin{align*}
P = & [0.5, 0.47, 0.01, 0.01, 0.01] \\
Q = & [0.96, 0.01, 0.01, 0.01, 0.01]
\end{align*}

$KL(P \lVert Q) = 1.48$ and $KL(Q \lVert P) = 0.58$. When the summed
terms are weighted according to $P$ in $KL(P \lVert Q)$, the first two
terms get large weight, while only the first term gets large weight in
$KL(Q \lVert P)$. When both KL divergences are small, it is a good
indicator of similarity of the two distributions.

\section{Maximum Entropy for MCTS and Theorem Proving}
\label{sec:maxent_atp}

In this section we discuss the entropy of the inference policy and its
potential influence on the proof search. We also introduce a
\emph{normalized entropy} to correctly handle probability vectors of
different lengths and argue that MaxEnt RL is more targeted and
powerful than previously used temperature-based entropy control.
Section~\ref{implementation} then describes our implementation.

\subsection{Exploration and Entropy in MCTS}

The MCTS implemented via the UCT formula has a built-in mechanism for balancing the
\emph{exploitation} of proof states that already have high rewards and
the \emph{exploration} of inferences whose effect is not yet known.
This balancing serves to mitigate errors in the proof state value
estimates. However, we need to do another balancing within
exploration, between the different under-explored inference branches. This is
estimated by the $\pi^M$ policy predictor as the prior probabilities of the possible inferences.  Hence,
besides the ordering of the inferences that are possible from a given state, their exact prior probabilities are important
as they determine how the exploration budget is split between them. This observation directs our attention to the entropy
(uncertainty) of $\pi^M$ and its relation to the theorem proving performance.

We argue that finding the right level of (un)certainty is particularly
important for theorem proving. The goal of learning is to acquire
inductive biases that allow the system to perform well on novel
problems.\footnote{In this sense, theorem proving can be considered as
  a meta learning task.} In many situations, however, it is not
realistic to extract enough knowledge from the training data that
justifies a high level of confidence. Sometimes, there is just not
enough analogy between a new problem and the training problems, and we
would like our guidance to be more conservative so that we can
explore all directions equally. This makes a strong case for using
MaxEnt RL, which gives tools for shaping the entropy (uncertainty) profile of our
learned $\pi^M$ policy predictor. 

\subsection{Normalized Entropy}
In this work, we empirically demonstrate the importance of
including the ``right amount'' of entropy when training the policy that guides the theorem prover.
To the best
of our knowledge, this is the first time that the effect of
entropy regularization for MCTS in general and for theorem proving in
particular is examined.

Using standard entropy for comparing probability vectors of different length
would, however, be misleading. The same entropy value can mean very different
uncertainty if the length of the vector changes. For example, consider
the vectors 

\begin{align*}
  & [0.34, 0.33, 0.33] \\
  & [0.73, 0.07, 0.05, 0.05, 0.05, 0.01, 0.01, 0.01, 0.01, 0.01] \\
\end{align*}
Their entropy is roughly the same (1.1), despite the fact that the
first is nearly uniform and the second centers around its first
value. To make the uncertainty of these two vectors comparable, we
introduce \emph{normalized entropy}:

\begin{definition}
Given a discrete probability vector $p$ of length $n$, let $H^*[p] =
H[p] / \log(n)$ denote the normalized entropy of $p$.
\end{definition}
Here, $\log(n)$ is the entropy of the uniform distribution when the length
is $n$, hence it is the upper bound of $H[p]$. Consequently, $H^*[p]
\in [0,1]$. Furthermore, it is dimensionless, i.e., it does not depend
on the base of the logarithm. The difference between the two
distributions in the example above is better captured by their
normalized entropy, which is 1 and 0.48.

\subsection{Temperature-based and Regularization-based Entropy Control}

An alternative mechanism for injecting entropy into the policy is
through the softmax \emph{temperature} parameter $T$. Our policy predictors (both
XGBoost and GNN) output an unconstrained \emph{logit} vector $l$,
which is normalized to a probability vector $p$ using the softmax
function:

$$p_i = \frac{e^\frac{l_i}{T}}{\sum_{j=1}^n e^\frac{l_j}{T}}$$
Increasing the temperature flattens the probability curve, approaching
the uniform distribution in the limit. On the other hand, if the
temperature gets close to $0$, then most of the probability mass
concentrates on the most likely action. 

While both higher temperature and entropy regularization
increase the ultimate entropy of the policy, they work differently. The
temperature acts globally and uniformly, flattening all inference probabilities
estimated  by the trained policy predictor. Entropy regularization, on the other hand, is
part of the training process and it allows the
neural network to learn distinguishing between situations with low and high uncertainty.
In
obvious situations, entropy regularization does not prevent the neural network
from acquiring great certainty, while it will drive the network to more uniform
predictions when there is no strong evidence against that in the
training data. Hence, we expect entropy regularization to be more
targeted and powerful than the temperature optimization. This is empirically
demonstrated in Section~\ref{sec:experiments}.

\section{Entropy Regularized Neural Guidance for \plc}
\label{implementation}

This section gives an overview of the training procedure, including how data is extracted from tableaux and Monte Carlo trees.

\subsection{Neural Representation of the State and Inference Steps}

The proof state in the \lc setting is roughly described by the partial
tableau and by the set of input clauses corresponding to the initial
axioms and conjecture. Each time we choose an extension step, we
map the state into a hypergraph, as described in
\cite{prop_invariant_embedding}. In more detail, we use the current
goal (single literal), the active path leading to the current goal (set of literals), the set
of all open goals (set of literals), and the input clauses in the hypergraph
construction. The GNN processes the hypergraph and outputs a value
prediction for the proof state in the range $[0,1]$. It also outputs a
probability distribution over all the literals of the axiom clauses
that can be used for extension steps, i.e., that can be unified with the
negation of the current goal literal.

The above method is used already in \gc, but the hypergraph construction algorithm used by \gc was designed to guide only the extension
steps. Hence it expects
the set of all
clauses together with the information that identifies literals
within the clauses that unify with the negation of the current
goal. We adapt this to paramodulation by temporarily creating a
clause for each valid paramodulation step that ``simulates'' the
latter as an extension step. Suppose that the current goal is $G$ and
there is an input clause $\{X \neq Y, B\}$, such that, for some position $p$
there is a substitution $\sigma$ such that $G|_p\sigma =
X\sigma$. There is a valid paramodulation step that replaces $G$
with $\{G[Y]_p\sigma, B \sigma\}$. We simulate this step as an
extension by adding clause $\{\lnot G\sigma, G[Y]_p\sigma, B \sigma\}$
to the input clauses, when constructing the graph.

\subsection{Training the Policy and Value Guidance for MCTS}

As in \plc, the value and policy estimates are stored in the MCTS nodes and are
used for guiding proof search. The training data for learning policy (inference probabilities) and value (state quality)
 are also handled as in \plc. They are extracted from the
tableau states of the bigstep nodes. For each bigstep state, the value
target is  1\footnote{A discount factor of 0.99 is applied to positive
  rewards to favor shorter proofs.} if it leads to a proof and 0
otherwise. The policy targets at a particular proof state
are the relative frequencies of the possible inferences, i.e., the children in the search tree. 

For \gc, a single GNN was jointly trained to predict
both the value and the policy~\cite{prop_invariant_embedding}. However, we observed that training
separate predictors yields a small improvement, hence we conduct our
experiments in this setup. Consider a tableau state $s$ for which we
want to learn its target value $v$ and target policy $p_1, \dots,
p_n$. Suppose that the partially trained value predictor outputs $v'$
and the policy predictor outputs $p'_1, \dots, p'_n$, then the objectives that we
minimize
are:
\begin{itemize}
  \item {\bf value objective}: $(v-v')^2$
  \item {\bf policy objective}: $- \sum_{i=1}^n p_i \cdot \log(p'_i) -
    \alpha H[p']$
\end{itemize}

For more details of the graph construction and neural training
we refer 
to \cite{prop_invariant_embedding}.
In summary, we use the same setting as there, except for (i) extending guidance to
paramodulation steps, (ii) training separate policy and value GNNs, (iii)
increasing the number of neural layers in the GNN from 5 to 10,\footnote{This is motivated by the experiments with the ENIGMA-GNN system~\cite{JakubuvCOP0U20}, where 8-10 layers produce better results than 5 layers.}
 and (iv) changing the policy training to encourage policies with higher entropy.


\section{Experiments}
\label{sec:experiments}

We first introduce our datasets and other experimental settings
(Section~\ref{datasets}). Then we show the impact of entropy
regularization (Section~\ref{influence}) and experimentally compare
the entropies and other characteristics of the XGBoost and GNN
predictors (Section~\ref{relative}).  In Section~\ref{order}, we show
that random policies with the right ordering and normalized entropy
perform already quite well. Section~\ref{temperature} then compares
temperature-based entropy control with our approach. Finally,
Section~\ref{final} evaluates the methods in a train/test scenario
on the full Mizar40 dataset using several iterations of proving and learning.

\subsection{Datasets and Common Settings}
\label{datasets}
We evaluate our system\footnote{The new extensions described here and the experimental configuration files are 
publicly available at \plc's repository: \url{https://github.com/zsoltzombori/plcop}.} using the same datasets as those in
\cite{rlcop}. The \emph{Mizar40} dataset~\cite{mizar40} consists of
32524 problems from the Mizar Mathematical Library that have been
proven by several state-of-the-art ATPs used with many strategies and
high time limits in the experiments described in
\cite{KaliszykU13b}. Based on the proofs, the axioms were
ATP-minimized, i.e., only those axioms were kept that were needed in
any of the ATP proofs found. 
The smaller \emph{M2k} dataset \cite{m2k} consists of 2003 Mizar40
problems that come from related Mizar articles.  Finally, we use the
bushy (small) problems from the MPTP2078
benchmark~\cite{abs-1108-3446}, which contains just an article-based
selection of Mizar problems, regardless of their solvability by a
particular ATP system.

Unless otherwise specified, we use the same hyperparameters as
described in \cite{plcop}, with the following important exceptions.
To allow for faster experiments and put more emphasis on guidance
instead of search, we reduce the per problem inference limit from
200000 to 20000 and the bigstep frequency from 2000 to 200. Hence the
overall search budget is reduced by a factor of 10. We use a very
large CPU time limit (300 sec) intended to ensure that proof search
terminates after exhausting the inference limit.

\subsection{Experiment 1: Influence of Entropy Regularization}
\label{influence}
In this experiment, we examine the effect of regularizing the entropy
of our policy predictor. We produce several variants of the GNN policy predictor which
differ in the entropy coefficient $\alpha$ used in its training. Table~\ref{tab:entropy}
summarizes our results. We find that the entropy coefficient has a big
impact on performance. By the 10th iteration, the best GNN predictor with
$\alpha = 0.7$ is $17\%$ better than the unregularized GNN and $5\%$
better than XGBoost. Table~\ref{tab:entropy} also shows the average
entropy of the policies generated by the predictor during the proof search.
Note that the average entropy of the best predictor in most iterations is
reasonably close (often the closest) to the entropy of the XGBoost
predictor. This suggests that one key strength of the XGBoost predictor is
that it hits the ``right'' amount of entropy. Matching this entropy in
the GNN with adequate regularization allows for matching and even
surpassing XGBoost in performance.

\begin{table}[htb]
  \caption{Number of problems solved (Succ) and average policy entropy (Ent) on the
  M2k dataset. $\alpha$ is the entropy loss term coefficient. Best
  models are marked with \textbf{boldface}, best GNN models are \underline{underlined}.}
  \label{tab:entropy}
  \centering
  \begin{tabular}{ l l | l r  | l r  | l r  | l r  | l r  | l r}
    & & \multicolumn{2}{c|}{Iter 1} & \multicolumn{2}{c|}{Iter 2} & \multicolumn{2}{c|}{Iter 4} & \multicolumn{2}{c|}{Iter 6} & \multicolumn{2}{c|}{Iter 8} & \multicolumn{2}{c}{Iter 10} \\
    Model & $\alpha$ & Ent & Succ & Ent & Succ & Ent & Succ & Ent & Succ & Ent & Succ & Ent & Succ \\
    \toprule
XGB &     & 1.41 & \textbf{790} & 1.29 & \textbf{956} & 1.22 & \textbf{1061} & 1.19 & 1119 & 1.17 & 1147 & 1.14 & 1171\\
GNN & 0   & 0.91 & 746 & 0.56 & 850 & 0.37 & 938 & 0.34 & 992 & 0.31 & 1021 & 0.32 & 1050\\
GNN & 0.1 & 0.86 & \underline{787} & 0.6  & 867 & 0.43 & 933 & 0.37 & 996 & 0.37 & 1031 & 0.38 & 1070\\
GNN & 0.2 & 1.11 & 769 & 0.71 & 878 & 0.51 & 976 & 0.51 & 1045 & 0.49 & 1077 & 0.46 & 1114\\
GNN & 0.3 & 1.05 & 736 & 0.8  & 868 & 0.7 & 991 & 0.73 & 1071 & 0.69 & 1109 & 0.78 & 1170\\
GNN & 0.5 & 1.31 & 781 & 1.14 & 884 & 1.17 & 1015 & 1.13 & 1085 & 1.12 & 1144 & 1.06 & 1191\\
GNN & 0.6 & 1.37 & 759 & 1.25 & 889 & 1.26 & 1040 & 1.21 & 1098 & 1.18 & 1150 & 1.19 & 1197\\
GNN & 0.7 & 1.41 & 727 & 1.32 & 854 & 1.27 & \underline{1057} & 1.22 &
\underline{\bf 1132} & 1.24 & \underline{\bf 1184} & 1.2 &
\underline{\bf 1228}\\
GNN & 0.8 & 1.42 & 757 & 1.37 & \underline{912} & 1.35 & 1029 & 1.32 & 1079 & 1.29 & 1111 & 1.3 & 1144\\
GNN & 1.0 & 1.53 & 742 & 1.41 & 911 & 1.38 & 1032 & 1.35 & 1102 & 1.36 & 1144 & 1.35 & 1173\\
GNN & 2.0 & 1.59 & 725 & 1.57 & 782 & 1.53 & 894 & 1.5 & 1007 & 1.5 & 1047 & 1.5 & 1086\\
    \bottomrule
  \end{tabular}
\end{table}

\subsection{Experiment 2: Relative Entropy on the Same Proof States}
\label{relative}
Table~\ref{tab:entropy} reveals that there is a reasonable match in
average entropy between XGBoost policies and our best GNN
policies. Note, however, that this is in general measured on different
proof states as the policies themselves determine what proof states
the prover explores. To gain a deeper understanding, we create a
separate dataset of proof states and compare the different GNNs from
Experiment~1 with XGBoost on these proof states using the following
four metrics: 1) fraction of proof states where the two predictors
have the same most probable inference (Best), 2) fraction of proof
states where the two predictors yield the same inference ordering
(Order), and the average KL divergence (relative entropy) between the
predictors in both directions: 3) $KL(X \lVert G)$ and 4) $KL(G \lVert
X)$.\footnote{$X$ and $G$ stand for the probability distributions
  predicted by XGBoost and GNN, respectively.} We contrast these
metrics with the number of problems solved (Succ) by the corresponding
entropy regularized GNN.

We perform this comparison using two datasets. These are the set of
states visited by an unguided prover on the 1) M2k dataset and the 2)
MPTP2078 bushy benchmark. The first set is part of the training corpus,
while the second was never seen by the predictors
before. The results can be seen in Table~\ref{tab:kl}.

\begin{table}[htb]
  \caption{Comparing the differently entropy-regularized GNN predictors (G) with XGBoost (X) on two fixed sets of
    proof states generated by running an unguided prover on the M2k and MPTP2078 benchmarks. All predictors
    were trained on M2k for 10 iterations. $\alpha$ is the entropy regularization coefficient. XGBoost solves 1171 (M2K) and 491
    (MPTP2078) problems. }
  \label{tab:kl}
  \centering
  \begin{tabular}{ c | c c c c c | c c c c c}
    & \multicolumn{5}{c|}{M2K} & \multicolumn{5}{c}{MPTP2078b} \\
    $\alpha$ & Succ & Best & Order & $KL(X \lVert G)$ & $KL(G \lVert
    X)$ & Succ & Best & Order & $KL(X \lVert G)$ & $KL(G \lVert X)$ \\
    \toprule
 0 & 1050 & 0.81 & 0.43 & 0.52 & 2.9 & 230 & 0.56 & 0.22 & 0.97 & 4.5\\
0.1& 1070 & 0.8  & 0.44 & 0.5  & 2.37& 245 & 0.58 & 0.24 & 0.91 & 3.83\\
0.2& 1114 & 0.81 & 0.42 & 0.47 & 1.66& 256 & 0.56 & 0.24 & 0.88 & 2.82\\
0.3& 1170 & 0.82 & 0.42 & 0.36 & 0.58& 276 & 0.56 & 0.23 & 0.61 & 0.9\\
0.5& 1191 & 0.82 & 0.42 & 0.24 & 0.28& 335 & 0.59 & 0.23 & 0.41 & 0.43\\
0.6& 1197 & 0.82 & 0.4  & 0.22 & 0.23& 359 & 0.59 & 0.23 & 0.36 & 0.36\\
0.7& \textbf{1228} & 0.82 & 0.4  & 0.22 & 0.21& \textbf{399} & 0.58 & 0.22 & 0.34 & 0.32\\
0.8& 1144 & 0.81 & 0.39 & 0.22 & 0.21& 357 & 0.58 & 0.22 & 0.34 & 0.31\\
1.0& 1173 & 0.82 & 0.4  & 0.24 & 0.21& 363 & 0.58 & 0.22 & 0.33 & 0.29\\
2.0& 1086 & 0.81 & 0.39 & 0.34 & 0.26& 362 & 0.58 & 0.21 & 0.37 & 0.3\\

    \bottomrule
  \end{tabular}
\end{table}


Changing the entropy coefficient mostly does not change the order of actions,
as expected. For the two datasets, the GNN and the XGBoost predictors select the same best
inference in around $80\%$ and $58\%$ of the states and yield exactly the same
inference ordering in around $40\%$ and $22\%$ of the states. This
reveals a significant diversity among the two predictor families,
suggesting potential in combining them. We leave this direction for
future work.

We find that the same level of entropy regularization ($\alpha = 0.7$)
is the best when running both on the familiar (M2k) and the previously unseen (MPTP2078) dataset. This is
where the two directional KL divergences (relative entropies) roughly coincide and their
sum is roughly minimal. These results make a stronger case for the
hypothesis from Experiment~1, that the best GNN performance is
obtained when the policy distributions are statistically close to
those of the XGBoost predictor.

\subsection{Experiment 3: Order and Entropy Are Largely Sufficient}
\label{order}

Tables \ref{tab:entropy} and \ref{tab:kl} demonstrate the importance
of the entropy of the inference policy in ATP performance. To make this even
more apparent, we design an experiment in which we remove a large part
of the information contained in the inference policy, only preserving the inference
ordering and the normalized entropy. The top two lines of
Table~\ref{tab:entropyN} show the normalized entropy across iterations
of the XGBoost and GNN predictors. Note that it is very stable. We select
a target normalized entropy $H^*$ and for each length $l$ we generate
a fixed random discrete probability $p_l$ of length $l$ whose
normalized entropy is $H^*$. Finally, we run an MCTS evaluation in
which each time our policy predictor emits a probability vector of length $l$, we
replace it with $p_l$, permuted so that the ordering remains
the same as in the original policy. Table~\ref{tab:entropyN} shows the ATP
performance for the differently normalized entropy targets. 

We find that the performance of this 
predictor is surprisingly
good: its performance (1154) is only $1\%$ worse than XGBoost (1171),
$10\%$ better than unregularized GNN (1050) and $6\%$ worse than the
best GNN (1228). This suggests that the right inference ordering plus
the right amount of entropy capture most of the benefits of the
learned inference guidance.

\begin{table}[htb]
  \caption{Normalized entropy of the XGBoost and GNN predictors on M2k
    (top two rows) and number of problems solved by random policies constrained to have the same action ordering and fixed normalized entropy.}
  \label{tab:entropyN}
  \centering
  \begin{tabular}{ l | l l l l l l l l l l l}
    \hline
    Iteration & 0 & 1 & 2 & 3 & 4 & 5 & 6 & 7 & 8 & 9 & 10 \\
    \hline
    XGBoost $H^*$ & 1 & 0.73 & 0.71 & 0.69 & 0.68 & 0.67 & 0.67 & 0.67
    & 0.66 & 0.67 & 0.66 \\
    GNN ($\alpha=0.7$) $H^*$ & 1 & 0.83 & 0.79 & 0.8 & 0.79 & 0.79 &
    0.78 & 0.78 & 0.78 & 0.8 & 0.78 \\
    \hline
    GNN $H^*=0.6$ & 523 & 700 & 782 & 849 & 909 & 956 & 984 & 1019 &
    1037 & 1059 & {\bf 1083} \\
    GNN $H^*=0.7$ & 523 & 702 & 800 & 856 & 922 & 954 & 995 & 1040 &
    1077 & 1110 & {\bf 1129} \\
    GNN $H^*=0.8$ & 523 & 693 & 832 & 938 & 1023 & 1054 & 1086 & 1077 &
    1115 & 1129 & {\bf 1154} \\
    \hline
\end{tabular}
\end{table}

\subsection{Experiment 4: Temperature vs. Entropy Regularization}
\label{temperature}
As noted in Section~\ref{sec:maxent_atp}, tuning the softmax
temperature is an alternative to entropy regularization. For XGBoost,
the temperature was previously optimized to be $T=2$ and all reported experiments use
this number. For the GNN predictors, we used the default $T=1$. In
Table~\ref{tab:temperature}, we show how the ATP performance of the GNN
changes after it has been trained for 10 iterations on the M2k dataset
(without entropy regularization). Increasing the temperature brings
some improvement, however, this is much smaller than the benefit of
entropy regularization. This is true even if we take the best
temperature ($T=4$) and perform a full 10 iteration training with this
temperature, as shown in Table~\ref{tab:temperature2}. We obtain $3\%$
improvement via the temperature optimization, compared with $17\%$
improvement via the entropy regularization.  We conclude that the effect
of entropy regularization is much more refined and powerful than just
flattening the probability curve.

\begin{table}[htb]
  \caption{The effect of changing the temperature of an
    (unregularized) GNN predictor trained for 10 iterations on the M2k dataset on the number of problems solved.}
  \label{tab:temperature}
  \centering
  \begin{tabular}{ l | l l l l l l}
    \hline
    Model & $T=0.5$ & $T=1$ & $T=2$ & $T=3$ & $T=4$ & $T=5$ \\ 
    \hline 
    GNN & 1036 & 1050 & 1057 & 1066 & {\bf 1068} & 1061 \\
    \hline
\end{tabular}
\end{table}

\begin{table}[htb]
  \caption{Number of problems solved by the GNN trained for 10 iterations on the M2k
    dataset with different softmax temperatures.}
  \label{tab:temperature2}
  \centering
  \begin{tabular}{ l | l l l l l l l l l l l l}
    \hline
    Iteration & 0 & 1 & 2 & 3 & 4 & 5 & 6 & 7 & 8 & 9 & 10 \\
    \hline 
    GNN $T=1$ & 523 & 746 & 850 & 899 & 938 & 971 & 992 & 1012 & 1021
    & 1023 & {\bf 1050}\\
    GNN $T=4$ & 523 & 705 & 800 & 864 & 894 & 931 & 993 & 1017 & 1049
    & 1065 & {\bf 1079}\\
    \hline
\end{tabular}
\end{table}

\subsection{Experiment 5: Final Large Train/Test Evaluation on Mizar40}
\label{final}

Finally, we perform a large evaluation of \plc using XGBoost and GNN
on the full Mizar40 dataset, and we compare its performance with \rlc
and \gc. This evaluation, including the training of the GNNs on the
growing sets of proofs generated from the successive proving/learning
iterations, takes over 10 days on a large multicore server, and the number
of training policy/value examples extracted from the MCTS proof searches goes over 5M in the
last iteration.

The 32524 problems are randomly split using a 9:1 ratio into 29272
training problems and 3252 evaluation problems. For consistency, we
employ the same split that was used in \cite{rlcop}. Successive
predictors are only trained on data extracted from the training
problems, but we always evaluate the predictors on both the training
and evaluation sets and report the number of proofs found for
them. Our results can be found in Table~\ref{tab:mizar40}, with
\plc/GNN solving in the last iteration 16906 training problems and
1767 evaluation problems. These are the highest numbers obtained so
far with any learning-guided \lc-based system.

For
training the GNN policy predictor, we use the entropy regularization
coefficient ($\alpha = 0.7$) that worked best on M2k, without its further
tuning on this larger dataset. Note that the resource limits are
higher in Table~\ref{tab:mizar40} for \rlc and also a bit different
for \gc (which was run only for a few iterations), as we took their
published results from \cite{rlcop,prop_invariant_embedding} rather
than rerunning the systems.  Also, the evaluation of \plc with GNN was
stopped after iteration 8 due to our resource limits and the clear
flattening of the performance on the evaluation set (1767 vs 1758 in the 8th vs 7th iteration).

\begin{table}[htb]
  \caption{Comparing \plc with XGBoost, \plc with GNN, \rlc and \gc on
    the Mizar40 training and evaluation set. \rlc employs 200000
    inference limit and 2000 bigstep frequency, \plc uses 20000
    inference limit and 200 bigstep frequency, \gc uses 200 depth
    limit and 200 bigstep frequency.}
  \label{tab:mizar40}
  \centering
  \begin{tabular}{ l | l l l l l l l l l l l}
    \hline
    Iteration & 0 & 1 & 2 & 3 & 4 & 5 & 6 & 7 & 8 & 9 & 10 \\
    \hline
        {\bf Train set} \\
        
        \rlc & 7348 & 12325 & 13749 & 14155 & 14363 & 14403 & 14431 &
        14342 & {\bf 14498} & 14481 & 14487 \\
        
        \gc & 4595 & 11978 & {\bf 12648} & 12642 \\

        \plc XGB & 4904 & 8917 & 10600 & 11221 & 11536 & 11627 & 11938 & 11999 & 12085 & 12063 & {\bf 12151} \\

        \plc GNN & 4888 & 8704 & 12630 & 14566 & 15449 & 16002 & 16467 & 16745 & {\bf 16906}\\
        \hline

        {\bf Eval set} \\
        \rlc & 804 & 1354 & 1519 & 1566 & 1595 & {\bf 1624} & 1586 & 1582 & 1591 & 1577 & 1621 \\

        \gc & 510 & 1322 & {\bf 1394} & 1360 \\

        \plc XGB & 554 & 947 & 1124 & 1158 & 1177 & 1204 & 1217 & 1210 & 1212 & {\bf 1213} & 1204 \\

        \plc GNN & 554 & 969 & 1375 & 1611 & 1650 & 1730 & 1742 & 1758 & {\bf 1767} \\
    \hline
\end{tabular}
\end{table}

In particular, \plc was given 20000 inferences per problem, i.e., one tenth of the
inference limit used for \rlc in~\cite{rlcop}. For a fair comparison with \rlc, we thus take the predictors (both XGBoost and GNN)
used in the last \plc iteration (iteration 10 for XGBoost and iteration 8 for GNN) and run \plc with them on the evaluation set with 200000
inference limit and 2000 bigstep frequency, which corresponds to the limits used for \rlc in~\cite{rlcop}.
To ensure that the system
has enough time to exhaust its inference limit, we increase the
timeout to 6000 seconds. \plc with XGBoost then solves 1499 of the evaluation problems while \plc with
GNN solves {\bf 1907} (58.6\%) of them. This is our final evaluation result, which
is {\bf 17.4\%} higher than the 1624 evaluation problems solved by \rlc in the best previously published result so far~\cite{rlcop}.

\section{Conclusion}

We have extended the \plc learning-based connection prover with a fast,
logic-aware graph neural network (GNN) and explored how the GNN can learn good guidance for
selecting inferences in this setting. We have identified the
entropy of the inference selection predictor as a key driver of the ATP
performance and shown that Maximum Entropy Reinforcement Learning largely improves
the performance of the trained policy network, outperforming simpler
temperature-based entropy increasing methods.
To the best of our knowledge, this is the first time that
the role of entropy in guiding a theorem prover has been 
analyzed.

We have discovered that replacing the particular trained predictors
by arbitrary (random) but entropy-normalized predictors that respect the inference ordering 
yields only slightly weaker theorem proving
performance than the best methods. This suggests that the right inference ordering plus the
right amount of entropy capture most of the benefits of the learned inference guidance.
In the large final train/test evaluation on the full Mizar40 benchmark
our system improves by $17.4\%$ over the 
best previously published result achieved by \rlc.

\section{Acknowledgments}
  ZZ was
    supported by the European Union, co-financed by the European
    Social Fund (EFOP-3.6.3-VEKOP-16-2017-00002), the Hungarian
    National Excellence Grant 2018-1.2.1-NKP-00008 and by the
    Hungarian Ministry of Innovation and Technology NRDI Office within
    the framework of the Artificial Intelligence National Laboratory
    Program. JU was funded by the \textit{AI4REASON} ERC Consolidator grant nr.
    649043 and the European Regional Development Fund under the Czech project AI\&Reasoning CZ.02.1.01/0.0/0.0/15\_003/0000466.
    MO was supported by the ERC starting grant no. 714034 SMART. We thank the TABLEAUX'21 reviewers for their thoughtful reviews and comments.

\bibliographystyle{abbrv}
\bibliography{plcop,ate11}

\appendix

\end{document}